\documentclass[11pt,a4paper]{article}
\usepackage[utf8]{inputenc}
\usepackage[T1]{fontenc}
\usepackage{lmodern}
\usepackage[margin=1in]{geometry}
\usepackage{microtype}
\usepackage{amsmath,amssymb,amsthm,mathtools}
\usepackage{graphicx}
\usepackage{booktabs,longtable,array,multirow}
\usepackage{enumitem}
\usepackage{caption}
\usepackage{xcolor}
\usepackage{tikz}
\usetikzlibrary{arrows.meta,positioning,fit}
\usepackage[colorlinks=true,citecolor=blue,linkcolor=blue,urlcolor=blue]{hyperref}
\usepackage{cleveref}
\usepackage{authblk}

\newcommand{\doi}[1]{\href{https://doi.org/#1}{\nolinkurl{doi:#1}}}

\newcommand{\Rsq}{R^2}

\title{Evolutional Math: symbolic regression for exact closed forms, with graded verification and reproducible
discovery records}
\author[1]{Artem Andrianov\thanks{Corresponding author: \texttt{artem.andrianov@evomath.ai}}}
\affil[1]{EvoMath Research Labs UG, M{\"u}nsterer Str.~49, 65719 Hofheim am Taunus, Germany}
\date{\today}

\begin{document}
\maketitle

\begin{abstract}
Evolutional Math (EvoMath) is a symbolic-regression system for one narrow task: recovering exact closed forms of
mathematical and physical quantities from numerical tables, and recording how each was checked. It combines
island-model genetic programming with exact-identification seeds, deterministic searches that propose closed forms
before evolution starts: least squares over libraries of special-function features (link functions, Laurent
polynomials, complete elliptic integrals, Gauss hypergeometric and Clausen functions), null-space fits of implicit
polynomial laws, and integer-relation searches over families of constants. Results are graded by the strength of the
evidence: numerical agreement on held-out points, symbolic identity after constant snapping, and derivations checked
step by step. A validity audit tests parameter domains, branch cuts, singularities, limits and dimensions, and a
discovery protocol ends in a sealed record of the starting point, data, search configuration, random seeds,
candidates, verification scripts and proof status. On 216 benchmark laws, with the same rows, budget (four
threads, 300 s per law) and symbolic judge, EvoMath recovered 179 laws exactly (147 in a first run, before two rounds
of changes prompted by its misses on these laws), against 125 for PySR, 60 for Operon and 27 for gplearn; it fitted
206 laws, PySR and Operon 194 and 193, and its median time was 8 s per law. Twenty-one constants given to 60 to 300
digits were all classified correctly: 18 identified and 3 controls refused. We also report the limits: the closed forms
behind five companion papers were derived first and only re-found by EvoMath, mostly with derived quantities supplied
as inputs or with seeds added after a miss. This version replaces the illustrative results of version 1 with
measurements. The code is at \url{https://gitlab.com/evomath-group/evomath}.

\end{abstract}

\section{Introduction}\label{sec:intro}

Symbolic regression searches for a formula that reproduces a table of numbers \cite{koza1992gp,schmidt2009eureqa}.
In most applications a short formula with a small error is the goal. In mathematics and theoretical physics the goal
is often stricter. A quantity is defined by an integral, a series or a boundary-value problem; it can be computed to
many digits; and the question is whether it has a closed form in known functions. Then a formula that agrees to six
digits is not an answer, and a formula that agrees to forty digits is a conjecture until it is proved. Integer-relation
algorithms such as PSLQ \cite{ferguson1999pslq,bailey2001parallel} have long served this purpose for constants
\cite{borwein2008experiment,raayoni2021ramanujan}. For functions of one or more parameters, symbolic regression is the
natural tool, but its usual output, a best-fitting expression with a goodness-of-fit score, does not say which of
three different things has been shown: that the formula agrees with the data, that it is identical to a known
expression, or that it is true.

Evolutional Math (EvoMath) is a symbolic-regression system built for this setting. The first version of this report
(May 2026) described its evolutionary core. Its experimental table was illustrative and its scaling figure was not
produced; this version replaces both with measurements, and describes the parts of the system that were built since:
\begin{itemize}[leftmargin=1.5em,itemsep=1pt]
\item \emph{exact-identification seeds} (\cref{sec:seeds}): deterministic searches, run before evolution starts, that
propose closed forms from libraries of special functions (by least squares), from implicit polynomial relations (by
null-space fits) and from families of constants (by integer-relation detection), and inject them into the populations
as ordinary candidates;
\item \emph{graded verification} (\cref{sec:verdict,sec:prooflevels}): a recovery verdict that separates symbolic
identity from numerical agreement, and a proof status that separates numerical evidence from checked derivations;
\item a \emph{validity audit} (\cref{sec:audit}) that tests a candidate formula for parameter domains, branch cuts,
singularities, limiting behaviour and dimensional consistency;
\item a \emph{discovery protocol} and a sealed \emph{discovery record} (\cref{sec:protocol,sec:record}) that keep the
analysis that motivates a search apart from the search itself, and preserve everything needed to repeat a result.
\end{itemize}
We compare EvoMath with PySR \cite{cranmer2023pysr}, Operon \cite{burlacu2020operon} and gplearn
\cite{stephens_gplearn} on 216 laws under one budget and one judge (\cref{sec:compare}), report constant
identification (\cref{sec:constants}), and give an account of what the system did, and did not do, in the mathematical
work that produced five companion papers (\cref{sec:cases}). That account is deliberately conservative. Every closed
form in those papers was found first by derivation, computer algebra or PSLQ; EvoMath re-found some of them from
numerical data, usually with derived quantities supplied as input columns or with a seed module added after a miss.
Under the protocol of \cref{sec:protocol} no result has yet reached the level of an EvoMath discovery.

\section{Related work}\label{sec:related}

\paragraph{Symbolic regression.}
Genetic programming over expression trees \cite{koza1992gp,poli2008field} remains the basis of most symbolic
regression systems (see \cite{kronberger2024book} for a recent account), with island models \cite{tomassini2005}, linear scaling \cite{keijzer2003scaling},
lexicase selection \cite{lacava2016epsilon} and nonlinear least squares for constants \cite{kommenda2020}. PySR
\cite{cranmer2023pysr} and Operon \cite{burlacu2020operon} are the strongest open implementations in the
benchmark of La Cava et al.\ \cite{lacava2021srbench}; gplearn \cite{stephens_gplearn} is a widely used baseline.
Other families use neural guidance \cite{udrescu2020aifeynman,udrescu2020aifeynman2,petersen2021dsr,
landajuela2022udsr,biggio2021nesymres,kamienny2022e2e}, Bayesian model selection \cite{guimera2020bms} or exhaustive
enumeration with a description-length criterion \cite{bartlett2023esr}. Symbolic regression is NP-hard in general
\cite{virgolin2022nphard}. Matsubara et al.\ \cite{matsubara2022srsd} argued that benchmarks for scientific discovery
should judge recovery by symbolic equivalence rather than by $\Rsq$, which is the criterion used here.

\paragraph{Units, domains and background knowledge.}
Dimensional analysis has been built into the search by AI~Feynman \cite{udrescu2020aifeynman} and by $\Phi$-SO
\cite{tenachi2023phiso}. AI-Descartes \cite{cornelio2023aidescartes} and AI Hilbert \cite{corywright2024aihilbert}
combine regression with background theory and check whether a candidate can be derived from axioms; this is the
closest prior work to a proof-aware pipeline. EvoMath does not derive candidates from axioms. Its proof levels
(\cref{sec:prooflevels}) record checked derivations written after the search.

\paragraph{Experimental mathematics.}
Integer-relation detection identifies constants from their digits \cite{ferguson1999pslq,bailey2001parallel,
borwein2008experiment}. The Ramanujan Machine \cite{raayoni2021ramanujan} searched for continued-fraction
conjectures for constants, and FunSearch \cite{romeraparedes2024funsearch} used program search with language models to
find constructions in combinatorics. EvoMath's constant seeds (\cref{sec:seeds}) are PSLQ searches over fixed
families of constants; its function seeds are least-squares searches over fixed libraries.

\section{The search engine}\label{sec:engine}

\Cref{fig:pipeline} shows how a run proceeds. This section describes the evolutionary core; \cref{sec:seeds}
describes the seeds and \cref{sec:verification} what happens to the result. Defaults are those of the configuration
object \texttt{GPConfig}; the law suites of \cref{sec:experiments} override a few of them (\cref{tab:defaults} in the
appendix).

\begin{figure}[!tbp]
\centering
\begin{tikzpicture}[node distance=4mm and 6mm, font=\small,
  box/.style={draw, rounded corners=2pt, align=center, inner sep=4pt, minimum height=9mm},
  arr/.style={-{Stealth[length=2mm]}}]
\node[box] (data) {data $(X,y)$\\train / holdout};
\node[box, right=of data] (seeds) {exact-identification\\seeds (\S\ref{sec:seeds})};
\node[box, right=of seeds] (gp) {island GP: lexicase,\\variation, simplification,\\constant optimisation};
\node[box, right=of gp] (arch) {archive and\\final choice\\(holdout, MDL)};
\node[box, below=9mm of arch] (verd) {recovery verdict\\exact / equivalent / none\\(\S\ref{sec:verdict})};
\node[box, left=of verd] (audit) {validity audit\\(\S\ref{sec:audit})};
\node[box, left=of audit] (proof) {proof status\\(\S\ref{sec:prooflevels})};
\node[box, left=of proof] (rec) {sealed record\\(\S\ref{sec:record})};
\draw[arr] (data) -- (seeds);
\draw[arr] (seeds) -- (gp);
\draw[arr] (gp) -- (arch);
\draw[arr] (arch) -- (verd);
\draw[arr] (verd) -- (audit);
\draw[arr] (audit) -- (proof);
\draw[arr] (proof) -- (rec);
\draw[arr] (data.south) |- ++(0,-4mm) -| ([xshift=-6mm]gp.south) ;
\end{tikzpicture}
\caption{A run. The seeds are computed once per island from the training rows and enter the initial populations as
ordinary candidates. The lower row applies to discovery dossiers; on benchmark laws only the recovery verdict is
computed.}
\label{fig:pipeline}
\end{figure}

\subsection{Representation and evaluation}\label{sec:repr}

A candidate is an expression tree of constants, input variables, unary and binary operators and, when granted, sums
over a bound index (finite, up to eight terms, or infinite, summed until three consecutive terms fall below $2^{-60}$
of the partial sum). The default operator set has 27 operators: $+$, $-$, $\times$, $\div$, power, minimum and maximum,
and negation, reciprocal, absolute value, square, cube, square and cube roots, $\log$, $\log_2$, $\log_{10}$, $\exp$,
$\sin$, $\cos$, $\tan$, $\tanh$, rounding, floor, ceiling, sign and the logistic function. Further operators are
available on request: $\sinh$, $\cosh$, $\log\Gamma$, $\operatorname{erf}$, the factorial, the generalised harmonic
number $H^{(b)}_a$, Clausen's function $\operatorname{Cl}_2$, the inverse trigonometric functions, Lambert's $W_0$, the
modular function $\lambda(q)=(\theta_2(q)/\theta_3(q))^4$ of the nome, and a family ${}_2F_1[a,b,c](z)$ with one member
per rational parameter triple, which enters only through a seed and is never used by mutation.

Evaluation is vectorised in float64 with the protective conventions usual in genetic programming
\cite{poli2008field}: a divisor below $10^{-10}$ in magnitude is replaced by 1, logarithms take $\max(|a|,10^{-10})$,
square roots take $|a|$, power exponents are clipped to $[-10,10]$, the argument of $\exp$ to $[-500,500]$, and every
value to $\pm10^{15}$. A formula whose evaluation reaches that clip receives a fixed penalty fitness and is never
archived. These conventions keep the search running; they are also the reason why a final formula has to be checked
again as ordinary mathematics (\cref{sec:audit}).

\subsection{Populations, islands and variation}\label{sec:islands}

Four islands evolve in parallel threads. Each has its own random stream, derived from one run seed, and an operator
subset (algebraic, logarithmic--exponential, trigonometric, full) intersected with the operators the user allows; an
island falls back to the full set when the intersection has fewer than five operators. Every 25 generations each
island sends its three best candidates to the next island in a ring, where they replace the worst. The islands share
one fitness cache and one set of formulas already seen, merged in island order after each generation so that a run is
repeatable for a given seed.

The initial population consists of structural templates (exponential decay, saturation, logarithmic and periodic
forms, rational forms, products of powers and others; at most half of the population), simple transforms and pairwise
combinations of the inputs, the seeds of \cref{sec:seeds}, and random trees. Parents are chosen by
$\varepsilon$-lexicase selection \cite{lacava2016epsilon} on 128 sampled training cases, with $\varepsilon$ the median
absolute deviation of each case's errors over the population; tournament selection is optional. Children are produced
by subtree crossover (probability 0.9) and mutation (probability 0.1: subtree 40\%, operator 20\%, Gaussian change of
the constants 20\%, hoist 20\%), then simplified by a rule-based pass (constant folding, collection of like terms,
algebraic identities). A child whose simplified form was evaluated before is mutated again or replaced, so no formula
is evaluated twice in a run. Three random trees enter each generation, and every 50 generations 200 random trees are
screened.

\subsection{Fitness, constants and model choice}\label{sec:fitness}

Fitness is the three-fold cross-validated $\Rsq$ of the linearly scaled model $a+b\,f(\mathbf{x})$
\cite{keijzer2003scaling}, with $(a,b)$ fitted on the other folds, each fold's score floored at $-1$, minus $0.005$ per
node. Because of the scaling the fitness does not distinguish a formula from an affine image of it; the first version
of this report argued for $\Rsq$ over Pearson's correlation, an argument that applies to unscaled models only. Fitness
is computed on a fixed subsample of 200 training rows; 15\% of the rows are held out from the search.

Every three generations the constants of the best member of each of the five best distinct structures are optimised
with L-BFGS-B \cite{byrd1995lbfgsb} (at most 50 iterations, on the squared error after the affine fit), followed by
exact least-squares refits of additive coefficients and of the exponents of products of powers.

The archive keeps one entry per structure, with the constants replaced by a placeholder; entries are simplified on
entry and ranked by holdout $\Rsq$. Their complexity is a description length: the node count plus a cost for each
constant that is not simple (integers up to 1000, rationals with denominator at most 12, and named constants such as
$\pi$ cost nothing). The archive is pruned to the non-dominated set in holdout $\Rsq$ and description length, and a
second front keeps the best entry at each length. The final choice takes the archive, any boosted models and the
variants obtained by dropping one or two operands, and picks the most accurate on the holdout unless shorter
candidates are not significantly worse; when $1-\Rsq\le10^{-9}$ the choice minimises a two-part code length whose
error term is floored at the rounding level of the target's printed digits, so that formulas exact to rounding are
not separated by noise.

On noisy data two further steps act on the final choice. Among candidates whose errors lie within the noise band, one
whose residual shows no structure (no input, ordering or neighbourhood predicts it) is preferred. Then each fitted
constant of the chosen model, together with the scaling coefficients, is refitted with its standard error, and a
constant is replaced by the simplest exact value of a fixed ladder (zero, integers, rationals $p/q$ with $q\le12$,
rational multiples of $\pi^{\pm1}$ and $e$, short decimals) when that value lies within 2.5 standard errors and no
value at most as simple lies within 4; the snapped model is kept only if it still holds on the held-out rows. Without
this step a law found with the right structure but constants a standard error off, such as $0.10105$ for $1/10$, could
not be judged exact.

\subsection{Stopping, restarts and budgets}\label{sec:stopping}

A run stops early when the holdout $\Rsq$ of the best archived formula reaches a target (0.9999; 0.999 in the law
suites) and stops improving, or at once above $1-10^{-7}$; the second test runs right after seeding and in every
generation, so an exact answer from a seed ends the run before any breeding. Noisy data cannot reach a target near 1.
With noise-aware stopping, which the law suites use, a best formula with $1-\Rsq\le0.01$ whose residuals cannot be
predicted by a five-nearest-neighbour average on the standardized inputs (leave-one-out $\Rsq$ of that average below
zero; noise gives about $-1/5$) counts as having reached the target. Optionally, a stall rule restarts all islands
from fresh seeded populations while keeping the archive and the cache. After the main loop, unless the target is
already reached or the residuals are noise in the sense above, one round of residual boosting fits a second formula
to the residual of the first, with no restarts and at most as many generations as the main loop used, and keeps the
sum only if it improves the holdout fit. A wall-clock budget, when set, covers seeding, the main loop and the
boosting. Seeding may use at most a share of the time left (40\% by default; recoveries in a discovery dossier use all
of it), before the first seeding and before each restart; once that share is spent, the remaining seed stages are
skipped. Seeds that several islands need over the same data, such as the
sparse seeds, which use no operators, are computed once per run. The
reported formula writes $\exp(c\,u)$ with $c=\pm\ln k$, $k\le10$ an integer, as $k^{\pm u}$ when the predictions
agree to a relative $10^{-10}$.

\subsection{Evaluation back-ends and other modes}

Fitness evaluation uses a process pool with the data in shared memory and trees sent as prefix strings; the pool
lives as long as the process and serves every run in it; for small problems evaluation runs in the island threads. GPU and mixed CPU--GPU back-ends exist; on the small tables of this paper they
are slower than the CPU and were not used. Further modes reuse the same engine: ordinary differential equations by
gradient matching with validation by integration, coupled systems, time-series maps and trends, and a discrete mode in
which the target is a constant given to many digits (\cref{sec:constants}). When the user declares units for the
inputs, every candidate is checked by a unit calculus (sums must be homogeneous and arguments of transcendental
functions dimensionless) and rejected if it fails.

\section{Exact-identification seeds}\label{sec:seeds}

Random variation finds short laws reliably and long ones rarely. A law with four additive terms needs three
coefficients that linear scaling does not supply, and a law in complete elliptic integrals of a modulus that is a
function of the input is out of reach of mutation. EvoMath therefore runs deterministic searches before evolution
starts. Each proposes closed forms of a fixed class and writes them into the initial populations as expression trees.
They are proposals: they compete in selection and reach the archive like any other candidate. \Cref{tab:seeds} lists
them; this section describes the ideas.

\begin{table}[p]
\centering
\small
\caption{Seed modules. ``Default'' modules run in every regression; ``opt-in'' modules run when the user enables them.
Function seeds work on data $y(\mathbf{x})$; constant seeds work in the discrete mode, where $y=c\,x$ and $c$ is given
to $P\ge32$ digits.}
\label{tab:seeds}
\begin{tabular}{@{}llp{9.2cm}@{}}
\toprule
Module & Status & Class of closed forms and method \\
\midrule
\multicolumn{3}{@{}l}{\emph{Function seeds}}\\
sparse & default & up to four monomials of degree $\le3$ in the inputs and their reciprocals; least squares over
supports found by elimination, forward selection and exhaustive search \\
link & default & $y=g^{-1}\big(c_0+\sum_{k\le4}c_kf_k(\mathbf{x})\big)$ for nine links $g$, over a library of up to 400
columns (powers, logarithms, exponentials, trigonometric and threshold transforms, compounding factors, products);
rational laws linearised by their denominator; a second pass with $\exp(\pm x_i/x_j)$ when no seed is exact; a
scaled logistic link; Lambert-$W$ links (with that operator); coefficients snapped to rationals \\
Laurent & default & $\sum c\prod_i x_i^{e_i}$ with $|e_i|\le6$, up to 12 terms and three inputs, optionally with
logarithms of the inputs (as terms or as factors); quartic polynomial boxes for three inputs; if $y$ has none, the
same search on $y^2$, $\log y$ and $1/y$, inverted; coefficients snapped to rationals \\
box & default & implicit laws $P+yQ$, $P+y^2Q$, $Ay^2+By+C=0$ over all monomials of degree $\le6$ in up to six inputs
and lens columns (trigonometric, decays, growth); exact data: a sharp null vector, exact on every row; noisy data: a
sparse relation with small rational coefficients and unstructured residuals \\
basis & opt-in$^{a}$ & $\sum_j g_j(p)\,F_j$ over user-named feature columns $F_j$ (periods, elliptic integrals, angles)
and small families of coefficient functions of a parameter $p$; the smallest exact support by exhaustive search;
one denominator or a power prefactor; coefficients snapped to $r$, $r\pi$ or $r/\pi$ \\
elliptic features & opt-in & complete elliptic integrals $K(m),E(m)$, their products and complements, and the mixed set
$\{K,E,K^2,KE,E^2\}$, at ten modulus maps $m(x)$ of each input (and $\lambda(e^{-\pi ax})$); passed to the basis seed \\
${}_2F_1$ & opt-in & $\alpha+\beta x^k\,{}_2F_1(a,b;c;z)$ with $10\,448$ rational parameter triples,
$z\in\{x,x^2,1-x,1-x^2\}$ and $k\in\{0,\pm\tfrac12,\pm1,\pm2\}$; screened by correlation, refitted, accepted at the
rounding level of the data on held-out rows \\
algebraic & opt-in & implicit laws $P(x,y)=0$ from an integer null vector of the monomial matrix, solved for $y$ when the
degree in $y$ is at most 2 \\
\midrule
\multicolumn{3}{@{}l}{\emph{Constant seeds (PSLQ)}}\\
relation & default$^{b}$ & Bailey--Borwein--Plouffe-type series (432 families) and quadratic surds \\
polylog & opt-in & products of weight $\le5$ of polylogarithmic constants for the alphabets $\{0,\pm1\}$ and
$\{0,\pm1,\pm i\}$, times powers of $\pi$ (26 bases, 118 trials) \\
$\mathbb{Q}(\sqrt3)$ & opt-in & polylogarithms of weight $\le3$ at $S$-unit arguments of $\mathbb{Q}(\sqrt3)$ \\
orthoscheme & opt-in & volumes of spherical 3-orthoschemes in Coxeter's dilogarithm form (371 angle triples, four
scales) \\
CM periods & opt-in & Chowla--Selberg periods of ten imaginary quadratic fields: monomials, two-term forms and period
polynomials \\
elliptic & opt-in & $K(m)$, $E(m)$ and $\pi$ at 35 rational moduli; period Laurent polynomials in ${}_2F_1(\tfrac1r,1-\tfrac1r;1;\tfrac12)$ \\
\bottomrule
\end{tabular}

\smallskip
\raggedright\footnotesize $^{a}$ Also called by default from the link seed, with each input in turn as the parameter
and the other inputs as features, when no exact link seed was found. $^{b}$ Active only in the discrete mode, which is
off by default.
\end{table}

\paragraph{Least-squares seeds.} The sparse, link, Laurent and basis seeds share one pattern: build a library of
columns from the inputs, fit linear combinations on the training rows, prune each support while it stays exact, and
admit a candidate only when its out-of-sample $\Rsq$ on held-back rows reaches $0.9999$ (or the rounding level of the
data, for the exact seeds). Linearisation extends the reach: a link function turns $y=\exp(\cdot)$ or a logistic law
into a linear problem, and a rational law $y=N/D$ becomes linear after multiplication by $D$. On noisy data, a fit
below the bar is admitted only when its residual is indistinguishable from noise (no library column, ordering or
neighbourhood correlates with it). Coefficients are snapped to simple rationals, or to rational multiples of $\pi$, only
when the snapped formula remains exact.

\paragraph{Implicit seeds.} The box seed covers laws that are polynomial or rational in the inputs, or a square or
quadratic root of such, when they have too many terms or inputs for the sparse and Laurent seeds: an implicit law
$\sum_k y^kP_k(\mathbf{x})=0$ makes the matrix of columns $y^k\times$ monomial rank-deficient, and its null vector is
the law. On exact data the seed accepts a relation only if the monomials are independent on the rows, the null vector
is sharp, and the pruned and refitted relation, solved for $y$, reproduces every row to $10^{-9}$; a box too wide to
hold rows back is fitted on all rows and checked by its pruned relation. Lens columns extend the class to laws in which
a few transcendental features enter polynomially (the law of cosines, decays with input-dependent rates). Under noise
a dense null vector is no test, so a second search looks for small subsets of columns that hold a relation, refits it
on $y$ by Gauss--Newton, and accepts it only when every coefficient is pinned to an integer or a small rational and the
residual shows no structure. When the data carry six to twelve digits, the bars follow the data's own rounding, and a
residual with structure rejects a polynomial approximant that passes every tolerance test. Both searches are bounded
by counts of candidate boxes and lenses, never by wall-clock time, so a result does not depend on machine load.

\paragraph{Special-function seeds.} The basis seed solves the case that defeats screening: a law
$y=\sum_j g_j(p)F_j$ whose terms cancel, so that no subset correlates with $y$ on its own. It fits all products of a
small family of coefficient functions with the given features at once and searches for the smallest support that
remains exact, within a budget of $70\,000$ fitted supports. The elliptic-feature seed supplies the features itself,
$K$ and $E$ at moduli such as $x^2$, $1-x^2$, $x/(1+x)$ or $(1-x)/(1+x)$, so that a law in complete elliptic integrals
can be found from the physical parameter alone, provided its modulus is in the list. The ${}_2F_1$ seed searches a fixed
grid of Gauss hypergeometric functions directly.

\paragraph{Constant seeds and false positives.} In the discrete mode the target is a constant $c$ known to $P$ digits,
posed as $y=c\,x$. Each constant seed runs PSLQ \cite{ferguson1999pslq} on $[c,b_1,\dots,b_d]$ for a basis $b$ of a
fixed family; for ten or more numbers it uses a two-level multipair variant \cite{bailey2001parallel}. A relation found
by chance is the main risk, so every search obeys one rule: it runs on the first half of the digits, with a height bound
$M$ chosen so that $d\log_{10}(2M+1)+\log_{10}T+3$ does not exceed the digits searched ($T$ is the number of trials in
the family), and a relation is accepted only if it reproduces all $P$ digits. The relation is then written out as an
exact tree, with series for the polylogarithms and $\pi$, so that the candidate \emph{is} the constant. In this mode
$\Rsq$ cannot separate an exact formula from a lookalike, so the final choice ranks candidates by the number of digits
of $c$ they reproduce when summed exactly in mpmath.

\paragraph{Where the seeds came from.} Several opt-in seeds were written after the engine missed a specific target,
and their documentation names it: the ${}_2F_1$ seed after the Kerr shadow perimeter, the elliptic seed after Finch's
constant, the orthoscheme seed after a box integral, the elliptic-feature passes after flat-band and duct-flow laws
(\cref{sec:cases}). The box seed and the noisy-relation search were written after the laws that the engine still
missed on the benchmark of \cref{sec:compare} had been sorted into classes (\cref{sec:lessons}). A success on the target
that motivated a seed shows that the seed works, not that the engine generalises. Since October 2026 the protocol requires that such a change widen a class, with positive tests on other
members of the class and negative controls outside it (\cref{sec:protocol}).

\section{From a fitted formula to a verified result}\label{sec:verification}

A symbolic regression run ends with a formula that fits the data. For a scientific claim three different statements
can be made about that formula, and EvoMath keeps them apart:
\begin{enumerate}[leftmargin=1.5em,itemsep=1pt]
\item \emph{numerical agreement}: the formula reproduces the target values, to a stated number of digits, on points
the search did not see;
\item \emph{symbolic identity}: the formula, after its fitted floats are replaced by exact constants, is equal to a
reference expression, and a computer-algebra system shows that the difference simplifies to zero;
\item \emph{proof}: a derivation, step by step, leads from the definition of the target to the formula, and each
step is justified.
\end{enumerate}
The first two can be decided by a program. The third, in the present system, is a written derivation whose steps
are checked by a program; it is not a formal proof (\cref{sec:prooflevels}).

\subsection{Recovery verdicts on benchmark laws}\label{sec:verdict}

On a benchmark the reference law is known, and each run is given one of three verdicts by
\texttt{recovery\_verdict} (\texttt{engine/equivalence.py}). The model under test is the one the engine reports,
$y=a+b\,f(\mathbf{x})$ with the linear-scaling coefficients included \cite{keijzer2003scaling}.
\begin{description}[leftmargin=1.5em,itemsep=1pt,font=\normalfont\itshape]
\item[exact] The tree is translated to SymPy \cite{meurer2017sympy}, each fitted float is replaced by the simplest
exact constant within a relative tolerance that tightens with the constant's denominator (an integer, a rational with
denominator at most 64, a rational multiple of $\pi$ or $e$ with denominator at most 24, or a rational with
denominator at most 1000; a value that already is such a constant to $10^{-12}$ keeps it), and the difference to the
reference law is reduced to zero by expansion, cancellation or simplification, with expressions under a root factored
first, under a time limit. Zero is accepted for a fitted intercept only when it is negligible against the scale of the
target, and is then tried first.
\item[equivalent] The symbolic step did not close, but the unsnapped model agrees with the law, to the same
tolerance relative to the root mean square of the law, on 400 points drawn in a window that extends each input range
by eight times its width. A polynomial that imitates a law on the
training window separates from it there by orders of magnitude.
\item[none] Neither.
\end{description}
Before either check, the translation itself is tested: the SymPy expression must reproduce the engine's own
predictions on the sampled domain to a relative $10^{-6}$. A translation that does not is reported as unreliable,
since every later statement would be about a different expression. A failed symbolic check means ``not proved'',
never ``proved different''. The same functions judge the outputs of the other systems in \cref{sec:compare}.

\subsection{Proof status of a discovery}\label{sec:prooflevels}

Outside benchmarks there is no reference law. For a candidate closed form produced in a discovery run
(\cref{sec:protocol}) EvoMath computes a proof status from the evidence in the dossier, as the highest of the
following levels whose conditions hold:
\begin{enumerate}[leftmargin=1.5em,itemsep=1pt]
\setcounter{enumi}{-1}
\item \emph{none};
\item \emph{numerical}: on held-out points the snapped formula, with no unsnapped constant left, agrees with the
target to at least 25 significant digits;
\item \emph{numerical, two routes}: in addition, a second numerical method, different from the one that produced the
data, agrees to $10^{-20}$ or better;
\item \emph{derivation, numerically checked}: a written derivation exists whose steps are recorded in a ledger; each
step holds at seeded rational points to 40 digits, and the last step ends at the candidate;
\item \emph{derivation, symbolically checked}: in addition, every ledger step carries a SymPy proof of the
equality it asserts;
\item \emph{formal}: reserved. No proof assistant is connected, and no result in this paper has this status.
\end{enumerate}
The status names the next level and what it needs, so that a reader can see what has not been done.

\subsection{Validity audit}\label{sec:audit}

Agreement on a holdout shows that a formula is right at the tested points. An identity claims more, and float64
evaluation with the engine's protected operators (\cref{sec:repr}) can hide a formula that leaves a principal branch
or divides by zero between data rows. The audit (\texttt{discovery/validity.py}) therefore evaluates the snapped
formula as ordinary mathematics in mpmath \cite{mpmath} against declarations made before the search, in the problem
statement (the domain and units of every parameter and of the target) and in the structural analysis (limits with an
error order, singularities by kind, parity). It checks:
\begin{description}[leftmargin=1.5em,itemsep=1pt,font=\normalfont\itshape]
\item[parameter domain] the data and holdout points lie in the declared domain, and the formula is real and finite at
Chebyshev interior points and on geometric approaches, $10^{-2}$ to $10^{-10}$, to each end of the domain (a grid
for two inputs); a $0/0$ with equal one-sided limits is recorded as removable;
\item[branch cuts] for every node with a restricted argument, whether a sample leaves the principal branch: square
root and logarithm, a power with a non-integer exponent, ${}_2F_1$ (argument below 1, or equal to 1 when
$c-a-b>0$), log-gamma and tangent;
\item[singularities] interior poles and cut crossings, located by bisection along each input and classified as pole,
removable point or cut; in the present version every one that is not removable fails the audit; branch points at
the ends of the domain are reported, and each declared singularity is tested against its kind (logarithmic, a pole
or power of stated order, or finite);
\item[limits] each declared expansion is compared with the formula along $h=10^{-2},\dots,10^{-11}$ (or the
reciprocal toward infinity) at 60 digits, and the observed convergence order must reach the declared order within
$0.3$;
\item[parity] when declared;
\item[dimensions] the formula's tree is passed through the engine's unit checker:
sums must be homogeneous, arguments of transcendental functions dimensionless, and the result must carry the
target's units.
\end{description}

\subsection{Discovery protocol}\label{sec:protocol}

A claim that EvoMath found a formula requires that the formula was not given to it. Since October 2026 discovery
work follows a protocol enforced by a command-line tool (\texttt{python -m evolmath.discovery}). A hunt is a dossier
with seven stages: literature, problem statement, analytical exploration, numerical data, recovery by EvoMath,
independent verification and review. A stage is frozen (hashed and time-stamped) when its gate passes; a later stage
may use only frozen material, and an edit after freezing is detected. The separation is what makes a result the
engine's:
\begin{itemize}[leftmargin=1.5em,itemsep=1pt]
\item the analysis may derive limits, singularities, symmetries and the natural function class, but may not identify
the closed form (no integer-relation search, rational fitting or derivation of the final expression on the target);
if the form becomes evident anyway, this is declared and the claim is capped;
\item the data come from a direct numerical route (quadrature, series, eigenproblems), never from a conjectured form,
and the holdout points are never shown to the engine;
\item the recovery runs are queued by the tool from the frozen data and specifications, with a default-engine
control, and each run records the engine's fingerprint (git HEAD, diff and untracked files);
\item an engine change made after a miss must widen a class of formulas (with positive tests on other members of the
class and negative controls outside it) rather than encode the answer;
\item verification uses a numerical method different from the one that made the data, and the review is done by a
reviewer with no part in the earlier stages, who re-runs the scripts in a throw-away copy.
\end{itemize}
The tool assigns a claim level: \emph{discovery} (all gates pass, the analysis did not identify the form, and the
novelty search found no prior publication), \emph{discovery, novelty unconfirmed}, \emph{rediscovery} (prior art
found), \emph{confirmation} (the form was identified before recovery) or \emph{incomplete}; each is qualified as
reviewed or awaiting review.

\subsection{Discovery record}\label{sec:record}

The final command writes \texttt{RECORD.json} and a readable \texttt{RECORD.md} at the root of the dossier. The
record is meant to let a reader repeat the result without the author. It contains:
\begin{itemize}[leftmargin=1.5em,itemsep=1pt]
\item the mathematical starting point: the definition, parameters with domains and units, why the problem is open,
the success criterion, the declared structure, and the evaluator scripts with their hashes;
\item the input data: the data file and the holdout, with hashes, ranges, precision and the route that produced
them;
\item the search configuration: every run specification as queued, and each run's fully resolved configuration
(all fields, defaults included), engine fingerprint, environment and data hash;
\item the provenance of the feature library: the operators, the enabled seed modules with their tuning constants,
and the SHA-256 of every engine module;
\item the random seed of every run (all randomness in a run derives from one seed);
\item every exact candidate with its holdout digits, and the selected formula;
\item the verification scripts and their outputs with hashes, the ledger with its symbolic proofs, the validity
audit, the novelty verdict, the review and the reproduction report;
\item the proof status and the claim level.
\end{itemize}
The record is sealed by a SHA-256 over its canonical JSON form, and \texttt{verify-record} recomputes every file
hash and the seal, so that an edited, added or removed file is detected.

\section{Experiments}\label{sec:experiments}

\subsection{Benchmark laws}\label{sec:suites}

The law suites are synthetic tables generated from published formulas (\cref{tab:suites}). Inputs are drawn uniformly
or log-uniformly in stated ranges, 400 rows per law (600 for eight laws of v6), with seed 42; noise, where present, is
additive Gaussian with a standard deviation that is a stated fraction of the target's. Each law's reference formula is
written so that the symbolic checker can parse it. The suites were assembled while the engine was developed, which is a
threat to validity that we discuss in \cref{sec:threats}; v5 was written to test the link-seed families on laws they
had not been tuned on, and includes four controls outside every family.

\begin{table}[!tbp]
\centering
\small
\caption{Benchmark suites.}
\label{tab:suites}
\begin{tabular}{@{}lrllp{5.4cm}@{}}
\toprule
Suite & Laws & Inputs & Noise & Content \\
\midrule
v1 & 50 & 1--3 & none & landmark formulas of physics (21), medicine (15), astronomy (6), chemistry (5), biology (3) \\
v2 & 50 & 1--5 & none & further published laws: medicine (18), physics (18), mathematics (14) \\
v3 & 50 & 1--5 & 0--5\% on 35 laws & mathematics, biology and medicine, with noise built in \\
v5 & 16 & 1--4 & 1--2\% on 3 laws & held-out test of the link-seed families, with 4 controls \\
v6 & 50 & 1--6 & 1--2\% on 14 laws & harder laws: rational functions of exponentials, piecewise forms, variable
exponents, wide log-uniform ranges, one irrelevant input in 8 laws \\
ODE & 5 + 3 & -- & 0 or 3\% & single-state equations (cooling, logistic, elimination, Michaelis--Menten, infusion) and
three coupled systems (two-compartment pharmacokinetics, Lotka--Volterra, SIR) \\
time series & 20 & -- & process or observation & 12 maps, 4 trends with cycles, 4 sampled ODEs; last 25\% held out \\
\bottomrule
\end{tabular}
\end{table}

\subsection{Recovery in the standard configuration}\label{sec:standard}

\Cref{tab:standard} reports a run of all suites with the released code (commit \texttt{6ba0eb6}, 25 November 2026)
in the configuration used for the engine's own regression tests (population 300 per island, at most 120 generations, depth 7,
four islands, early stop at holdout $\Rsq=0.999$ with a stall rule and two restarts, seed 42, no time limit). ``Fit''
is holdout $\Rsq\ge0.999$; ``exact'' and ``equivalent'' are the verdicts of \cref{sec:verdict}, with tolerance
$\max(10^{-4},3\sigma)$ for noise level $\sigma$. The same configuration with a 300\,s limit, judged by the shared
judge of \cref{sec:compare}, is the second row of \cref{tab:compare}. The run shared the machine with other jobs, so its
times are not a measure of speed.

\begin{table}[!tbp]
\centering
\small
\caption{Standard configuration, released code, one seed. Recovered = exact + equivalent.}
\label{tab:standard}
\begin{tabular}{@{}lrrrrr@{}}
\toprule
Suite & Laws & Fit & Exact & Equivalent & Recovered \\
\midrule
v1 & 50 & 50 & 45 & 5 & 50 \\
v2 & 50 & 50 & 48 & 2 & 50 \\
v3 (noisy) & 50 & 44 & 42 & 6 & 48 \\
v5 & 16 & 16 & 15 & 1 & 16 \\
v6 & 50 & 41 & 31 & 6 & 37 \\
\midrule
total & 216 & 201 & 181 & 20 & 201 \\
\bottomrule
\end{tabular}
\end{table}

On the noiseless suites most fits are exact recoveries. Noise moves some recoveries from ``exact'' to ``equivalent'':
at 1--5\% noise a fitted constant is not always pinned to one simple value, but the model still agrees with the law
far outside the training ranges. The v6 laws are where the engine fails most: of its 13 non-recoveries, 9 do not reach
the fit threshold and 4 fit the data without being the law. The differential-equation benchmark recovered 5 of 5
single-state equations without noise and 5 of 5 at 3\% noise (by the benchmark's trajectory and gradient criteria), and
2 of 3 coupled systems. In the time-series suite 14 of 20 laws were recovered, 12 of them exactly. Two suites written
for opt-in operators, and run with them granted, gave 13 of 16 exact recoveries for the inverse trigonometric
functions and 9 of 10 for Lambert's $W$.

The same run with the code of 20 November 2026 (commit dca142b) recovered 154 laws exactly and 175 in all, and 145
and 170 with the development tree of 31 October, before the changes of \cref{sec:lessons}. The 27 additional exact
recoveries come from the changes of \cref{sec:round2}: 28 laws that were not recovered or only equivalent are now
exact, among them the law of cosines, the Biot--Savart field, the Hill equation, competitive inhibition and the
two-compartment bolus, and one exact law, the Kelly fraction, became equivalent. The time-series suite has the same 14
recoveries as on 20 November, and four more of them are exact. The differential-equation benchmark gave the same
results.

\subsection{Comparison with other systems}\label{sec:compare}

\paragraph{Protocol.} PySR \cite{cranmer2023pysr} (version 2.7.2, with Julia 1.11.9), Operon \cite{burlacu2020operon}
(pyoperon 0.6.1) and gplearn \cite{stephens_gplearn} (0.4.3) were run on the 216 laws of v1--v6 with the rows that
EvoMath sees, exported by EvoMath's own data generator. Every system had the same budget per law: four threads and a
wall-clock limit of 300\,s. Each was configured close to its recommended settings with a comparable operator set
(\cref{tab:competitor-settings} in the appendix): PySR with its default search, a maximum size of 30 and its recommended
constraint on powers; Operon with its default budget of 1000 generations and $10^6$ evaluations; gplearn with a
population of 1000, continued generation by generation until the time limit. PySR and gplearn stopped early when the
training error reached rounding level. EvoMath ran its suite configuration (which stops when the holdout $\Rsq$
reaches 0.999 and stops improving) with the same limit and four evaluation workers.

The final formula of every system was judged by EvoMath's verdict functions (\cref{sec:verdict}): the same constant
snapping, the same time-limited symbolic check and the same extrapolation probe. For the other systems the formula was
first exported as a SymPy expression and checked against the system's own predictions on the holdout (to a relative
$10^{-6}$; $10^{-4}$ for Operon, which evaluates in single precision); and every float
whose removal changed the model by less than the tolerance was set to zero, the counterpart of the zero test that
EvoMath applies to its own fitted intercepts. Holdout $\Rsq$ is computed on 400 noiseless rows drawn with a different
seed.

\begin{table}[!tbp]
\centering
\small
\caption{Comparison on the 216 laws under one budget (four threads, 300\,s per law, seed 42) and one judge. Each suite cell gives exact recoveries / all recoveries (exact or equivalent). Fit: holdout $R^2\ge0.999$ on noiseless rows. Times are median seconds per law.}
\label{tab:compare}
\setlength{\tabcolsep}{3.3pt}
\begin{tabular}{@{}lcccccrrrr@{}}
\toprule
System & v1 (50) & v2 (50) & v3 (50) & v5 (16) & v6 (50) & Exact & Recov. & Fit & Time \\
\midrule
EvoMath, full budget$^{a}$ & 44/50 & 48/50 & 42/48 & 15/16 & 30/37 & 179 & 201 & 206 & 8 \\
EvoMath, suite configuration$^{a}$ & 44/50 & 48/50 & 42/48 & 15/16 & 31/37 & 180 & 201 & 206 & 8 \\
EvoMath, code of 20 November$^{b}$ & 44/50 & 46/47 & 35/42 & 12/14 & 17/23 & 154 & 176 & 193 & 7 \\
EvoMath, first run$^{c}$ & 44/50 & 44/45 & 31/39 & 11/13 & 17/23 & 147 & 170 & 189 & 26 \\
\midrule
PySR$^{d}$ & 36/41 & 34/38 & 32/38 & 8/9 & 15/18 & 125 & 144 & 194 & 300 \\
Operon$^{d}$ & 31/43 & 18/30 & 8/19 & 3/6 & 0/7 & 60 & 105 & 193 & 16 \\
gplearn$^{d}$ & 14/14 & 9/9 & 3/4 & 0/1 & 1/1 & 27 & 29 & 39 & 301 \\
\bottomrule
\end{tabular}
\par\smallskip\raggedright\footnotesize $^{a}$ Released code (tag \texttt{evomath-paper-v2}, commit \texttt{6ba0eb6}, run on 25 November 2026), with the changes of \cref{sec:lessons,sec:round2}; ``suite configuration'' stops as in the engine's own benchmarks (120 generations, two stall restarts), ``full budget'' keeps searching until the limit or $1-R^2\le10^{-7}$. $^{b}$ Code of 20 November 2026 (commit \texttt{dca142b}), full budget, before the changes of \cref{sec:round2}. $^{c}$ Development tree of 4 November 2026, before the changes of \cref{sec:lessons}. $^{d}$ Runs of 4 November 2026. Every row was judged by the scorer of the released code.
\end{table}

\begin{figure}[!tbp]
\centering
\includegraphics[width=\textwidth]{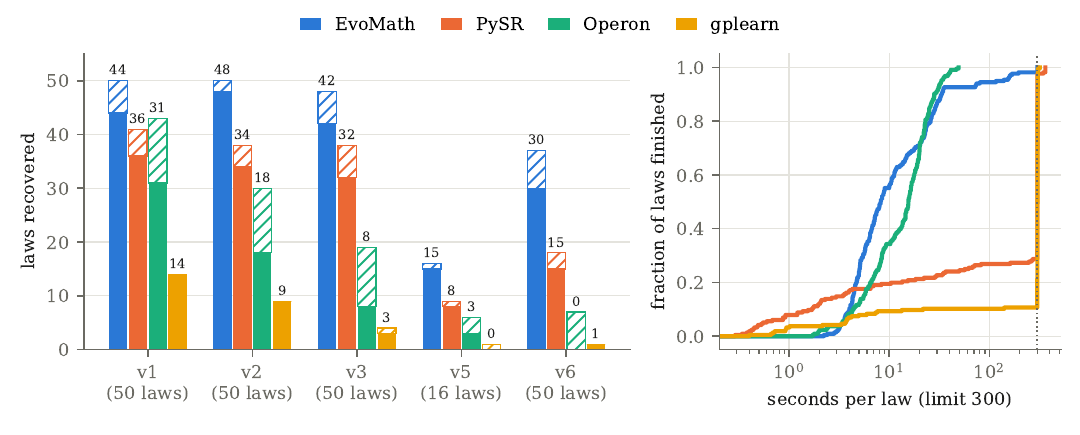}
\caption{Left: laws recovered per suite; solid bars are exact recoveries (the number above each bar), hatched
segments equivalent ones. Right: distribution of the time per law; the dotted line is the 300\,s limit. The data are in
\cref{tab:compare} and in the ancillary file \texttt{fig\_compare.csv}.}
\label{fig:compare}
\end{figure}

\paragraph{Results.} With the released code and the full budget, EvoMath recovered 179 of the 216 laws exactly and 201
exactly or equivalently, against 125 and 144 for PySR, 60 and 105 for Operon, and 27 and 29 for gplearn
(\cref{tab:compare}, \cref{fig:compare}). It recovered the most laws exactly in every suite; in v3, the suite with
noise, 42 against 32 for PySR. PySR recovered five laws exactly that EvoMath did not (the Black--Scholes $d_1$, mixed
enzyme inhibition, the Kelly fraction, log-sum-exp and serum osmolality; EvoMath recovered all of them but the first
as equivalent), and EvoMath 59 that PySR did not. EvoMath fitted 206 laws to holdout $R^2\ge0.999$, PySR 194 and
Operon 193. The gap between fitting a law and recovering it is widest for Operon, which fits 193 laws and recovers
105; on v6 its median formula has 51 operations and symbols, against 18 for PySR and 14 for EvoMath. EvoMath stopped
after a median of 8\,s per law and Operon ran its own budget in 16\,s; EvoMath used the whole 300\,s on 4 laws, all
unsolved, while PySR and gplearn stop early only on a near-exact fit and used the whole 300\,s on 154 and 193 laws. gplearn's protected division and
logarithm make 121 of its exported formulas differ from its own predictions; its verdicts are those of the exported
formulas.

\subsection{What the comparison showed: a first round of changes}\label{sec:lessons}

The first run (\cref{tab:compare}, fourth row) was analysed law by law. Where PySR or Operon recovered or fitted a law
and EvoMath did not, the laws fell into five groups. (i)~A polynomial under an outer function: the Gaussian density,
whose logarithm is a polynomial in $x$, $\mu$ and $1/s$ plus $\log s$; Heron's area, whose square is a quartic; the
law of cosines; the catenary $a\cosh(x/a)$. (ii)~Rational laws with several denominator terms, such as competitive and
mixed enzyme inhibition, which PySR writes directly. (iii)~Variable exponents, as in the Hill equation
$V_{\max}/(1+(K/S)^n)$. (iv)~Noise: the seeds admit only exact fits, or noisy fits with a white residual, so on noisy
laws the evolutionary search must find the form. (v)~Budget and throughput: in the suite configuration EvoMath stopped
after a median of 26\,s, while PySR and gplearn always spend the full 300\,s; with the whole budget it still completes
only about 100 generations of $4\times300$ candidates, far fewer evaluations than PySR.

Four changes followed, each aimed at a class of laws rather than at a law, each with positive tests on other members of
the class and negative controls: (1)~when $y$ has no exact Laurent law, the Laurent seed now searches $y^2$, $\log y$
and $1/y$ and inverts the result; (2)~logarithms of the inputs enter that search also as terms of their own, and three
inputs also get polynomial boxes of degree four; (3)~when the link seed finds no exact seed, a second pass adds the
columns $\exp(\pm x_i/x_j)$; (4)~the verdict now tries zero first for a fitted value that is negligible against the
target's scale (before, a noise-level intercept of 33.65 on a target of order $10^4$ snapped to 34, and the law was
judged equivalent rather than exact). In the suite configuration the changes added four exact recoveries and lost none
(151 exact and 174 in all, against 147 and 170): the Gaussian density and Heron's area through (1) and (2), and the
catenary and capacitor charging $V_0(1-e^{-t/\tau})$ through (3). The last was not a target of the change: it is one
of the four controls of v5, written to lie outside the link-seed families of that time. The differential-equation
and time-series benchmarks gave the same results as before. The full-budget configuration, which keeps searching
until the time limit or until $1-R^2\le10^{-7}$, is the fair setting for a time-capped comparison, since the other
systems spend their budget; with it the same engine recovered 151 laws exactly and 175 in all.

\paragraph{Time per law.} The time distribution of that full-budget run had two flat stretches: no law finished in
under about 9\,s, and none between about 35 and 110\,s. Both came from when a run stopped, not from the search. The
seeds found most laws before the first generation, but the target was tested only at every fifth generation, and
each run started a fresh pool of four worker processes. Of the 141 laws that reached the target, 140 reached it at
that first test, in 9 to 33\,s; uncontended, about half of such a run went to seeding and starting the pool, the other
half to the five generations before the test. The other 75 laws never reached $1-\Rsq\le10^{-7}$: all 52 noisy laws,
which cannot reach it even with the exact formula (36 of them were recovered), and 23 unsolved noiseless ones. In 64
of them the stall rule ended the main loop, at a median of 40 generations, and residual boosting then fitted the
residual, which on noisy laws is the noise, for up to 200 generations with about ten restarts: 86\% of these runs'
generations, for a better holdout fit in 15 of the 64, in the fourth to sixth decimal. Four further changes followed
(\cref{sec:stopping}): the exact-class test right after seeding and in every generation; one worker pool per process;
noise-aware stopping, which treats a fit whose residuals a nearest-neighbour average cannot predict as having reached
the target; and boosting skipped for such fits, and otherwise run without restarts and for at most as many
generations as the main loop. On the 4 November results the residual score separated the cases cleanly: $-0.30$ to
$-0.05$ for noisy laws whose law was recovered (one outlier), $+0.07$ to $+1$ for unsolved noiseless ones. A model-free
estimate of the noise variance (the Gamma test \cite{stefansson1997gamma}) was tried first and rejected: with a few
hundred rows over wide input ranges the nearest neighbours are far apart and curvature swamps the noise, and it
flagged 46 of 164 noiseless laws as noisy. A rerun with these changes lost the exact verdict on the Titius--Bode law:
stopped after seeding, it reported $\exp(0.693147180559945\,n)$ instead of $2^n$, a constant no verdict snaps. The
reported formula now writes $\exp(\pm\ln k\,u)$ as $k^{\pm u}$ for integers $k\le10$; this restored the law and made
the doubling law $N_0\,2^{t/T_d}$ exact, judged equivalent in every earlier run for the same reason.

With all these changes the full-budget run of 7 November took 2.2\,h of computing instead of 5.3\,h, the median time
per law fell from 19 to 8\,s, and 58\% of the laws finished within 10\,s. It recovered 154 laws exactly and 176 in
all. Five verdicts changed: three noisy laws whose longer runs had settled on lookalikes became exact (Winters'
formula, the Karvonen heart rate and the shunt fraction), the doubling law became exact, and mixed enzyme inhibition,
stopped at a fit within the noise, fell from exact to equivalent. In the suite configuration the time fell from 2.5
to 1.5\,h and the median from 18 to 7\,s; two laws became exact (the shunt fraction and the doubling law) and none were
lost. The differential-equation and time-series benchmarks gave the same counts as before; the clean ODE suite ran in
20\,s instead of 65\,s.

\paragraph{Code of 20 November.} The code of 20 November 2026 (commit dca142b) added one change to the engine of
7 November: seeding may use only a share of the remaining time (\cref{sec:stopping}), so that a run no longer overruns
its time limit while seeding; on 7 November the longest run took 323\,s. With it, every law gave the same verdict as on
7 November in both configurations. The full-budget run, the third row of \cref{tab:compare}, took 2.4\,h of computing;
the median time per law was 7\,s, and no run took longer than 302\,s.

These changes were motivated by failures on the same suites on which they were then measured, and the noise test's
threshold was read off these suites' residuals, so the counts after them overstate what the changes would gain on
unseen laws. The first-run row is kept for that reason. Groups (ii) to (v) remained open; the next section describes a
second round of changes aimed at them.

\subsection{A second round of changes}\label{sec:round2}

In the suite configuration the code of 20 November recovered 153 laws exactly and 175 in all. The laws it missed were
sorted again, and four changes followed, each aimed at a class, each with its own tests on planted laws, and each
admitted by a gate: a run of the 216 laws in the suite configuration, scored by the shared
judge, and the differential-equation, time-series, inverse-trigonometric and Lambert-$W$ suites. The counts below are
those gates, which ran from frozen copies of the development tree on 22--24 November 2026.
\begin{enumerate}[leftmargin=1.5em,itemsep=1pt]
\item \emph{Noise-band snapping} (\cref{sec:fitness}). On several noisy laws the engine had the law's structure with
constants about a standard error from simple values, such as $0.10105$ for $1/10$ in the $Q_{10}$ temperature law. The
final constants are now snapped when the data pin them to one simple value. The same cycle corrected the judge, which
under the noisy tolerance had re-snapped exact small values to coarse rationals (an exact $0.216$ to $8/37$). Exact
recoveries rose from 153 to 157, none were lost, and the time-series suite gained three exact recoveries.
\item \emph{A logistic link on a scaled input} (\cref{tab:seeds}): one law, island biogeography (158).
\item \emph{The box seed} (\cref{sec:seeds}). Of the 39 laws still not recovered, eight were exact polynomial, rational
or quadratic-root laws with four to six inputs or more than four terms, out of reach of the sparse and Laurent seeds,
and five were sums of exponentials with input-dependent rates. Exact recoveries rose from 158 to 173 and all
recoveries from 177 to 193, and no law was lost. Among the gains are the Biot--Savart field on the axis of a loop, the
law of cosines, the area of a triangle from its corners and the two-compartment bolus. For the Biot--Savart form the
judge gained one proof step, the factoring of a radicand; it changed no verdict of the other systems.
\item \emph{The noisy-relation search} (\cref{sec:seeds}). A probe in which the true law was given to the run as a seed
showed that on 13 of the 15 noisy laws then missed the law wins the final choice, so the gap was finding it. Exact
recoveries rose from 173 to 180 and all recoveries from 193 to 201, among them the Hill equation, competitive
inhibition and the nonrectangular photosynthesis law. One law was lost: the Kelly fraction, given its own law by a
seed, came back with the factor $e^{1/W}$, $W$ an input between 100 and 10\,000 that the law does not contain, in
place of its constant 1, and is now judged equivalent instead of exact.
\end{enumerate}
Other changes in the same days had no effect on these laws: the opt-in operators $W_0$ (with its own suite of ten
laws, nine exact with the operator and none without) and $\lambda(q)$, mixed-degree elliptic features, a preference for
unstructured residuals in the noise band, the removal of step operators (rounding, floor, sign) that are constant on
the data, and the seeding share as a setting. Two measurements suggest that more search
would not have helped: given 8 restarts and 400 generations, the 23 laws still missed after the box seed recovered none,
and with the early stop disabled, none of the four laws that stopped on a lookalike was recovered.

\paragraph{Released code.} The released code (commit \texttt{6ba0eb6}) contains both rounds. Both configurations were
run with it on 25 November 2026 from a clean export of that commit; these runs are the first two rows of
\cref{tab:compare} and the data of \cref{fig:compare}. In the suite configuration every law gave the same verdict as
in the last gate, and all but one the same formula: 180 exact recoveries and 201 in all. The full-budget run differs
from it on one verdict: the Katch--McArdle formula $370+21.6\,W(1-B/100)$, a noisy law, is exact in the suite
configuration, where noise-band snapping fixed its constants, and equivalent with the full budget, where the run ended
on the expanded form $21.59\,W-0.2159\,WB$, whose constants the snapping left as they were. The full-budget run took 1.4\,h of
computing (2.4\,h on 20 November, when more laws searched until the limit; timings on this shared machine also vary by
up to 30\% between days), the median time per law was 8\,s, 56\% of the laws finished within 10\,s, and the four runs
that reached the limit were laws that were not recovered.

Like the first round, these changes were made after the misses on these laws had been examined, and they were measured
on the same laws, so the first two rows of \cref{tab:compare} overstate what the engine would gain on unseen laws. Two
checks speak to false positives rather than to generalisation: on all 216 laws the noisy-relation search returned 46
relations, each the law and none a lookalike; and on the 164 noiseless laws rounded to six digits, where an
approximant passes every tolerance test, the box seed proposed no false law, while on 24 of them run end to end it
raised the exact recoveries from 11 to 17. The fifteen laws still missed include three with an input as an exponent
(the power mean, Richards growth and the real compound annual growth rate), three in which two kinds of feature enter
one function (the Black--Scholes $d_1$, the two-dimensional Gaussian density and the Mahalanobis distance), two noisy
laws with several terms (the Sampson LDL formula and Allee growth) and a piecewise law (CKD-EPI 2021). Of the open
groups of \cref{sec:lessons}, noisy rational laws (ii) and the Hill equation (iii) are now recovered; general variable
exponents and throughput (v) remain open.

\subsection{Constant identification}\label{sec:constants}

Twenty-one constants were given to 60--300 digits and posed in the discrete mode as $y=c\,x$, $x=1,\dots,200$, with all
constant seeds enabled. Eighteen have known closed forms in six families; three are controls that should not be
identified: Euler's constant $\gamma$, $e^\pi$ and a random number. A constant counts as identified when the returned
tree, summed exactly, reproduces all given digits but one. The released code identified all 18 and none of the controls
(\cref{tab:constants}), as the code of 20 November and the development tree of 16--17 October 2026 had done. The same benchmark identified 5 of the
18 at commit 4d035d4, before the surd and polylogarithmic seeds, and 9 at commit e3a0856, which added them. Three
families (the $\mathbb{Q}(\sqrt3)$, orthoscheme and CM-period seeds) were written during this campaign with these
constants in view, so for those families the table shows that the seeds work, not how they generalise. Every
identification came from a seed before the first generation and ended the run there (\cref{sec:stopping}), in 1--3
minutes per constant (3--13 minutes in October, when the target was first tested after five generations, except the
first October runs of the three orthoscheme constants, which took 74--121 minutes); the controls ran to the stall
limit (9--16 minutes).

\begin{table}[!tbp]
\centering
\small
\caption{Constant identification (released code, commit \texttt{6ba0eb6}).}
\label{tab:constants}
\begin{tabular}{@{}llr@{}}
\toprule
Family & Constants & Identified \\
\midrule
series & $\pi$, $\pi^2$, $\log2$, $\log^22$, $\pi\sqrt2$ & 5/5 \\
polylogarithmic & Catalan's constant, $\zeta(3)$ & 2/2 \\
surds & $\varphi$, $\sqrt{33}/4$ & 2/2 \\
$\mathbb{Q}(\sqrt3)$ polylogarithms & two cosmological phase constants & 2/2 \\
orthoscheme volumes & box-integral constants $J(2)$, $J(3)$, $B_6(-7)$ & 3/3 \\
CM periods & Finn--Chernoff moment, $K(1/\sqrt2)$, $\Gamma(1/3)$, $\Gamma(1/4)$ & 4/4 \\
\midrule
controls & $\gamma$, $e^\pi$, a random number & 0/3 \\
\bottomrule
\end{tabular}
\end{table}

\section{Case studies: what EvoMath did in the mathematical work}\label{sec:cases}

From October 2026 the system was used in about forty short campaigns on quantities from physics and applied
mathematics that were known only as integrals, series or tables. In each, the closed form was first obtained by other
means, by derivation, computer algebra (SymPy Hermite reduction), least squares on special-function bases or PSLQ on
the target, and EvoMath was then run on numerical data to see whether it would find the same law. In the terms of
\cref{sec:protocol} these are \emph{confirmations}. Five of the results are the subject of companion papers
\cite{andrianov2026kerr,andrianov2026lens,andrianov2026df,andrianov2026bsr,andrianov2026nfw}; \cref{tab:cases}
summarises what the engine contributed to each.

\begin{table}[p]
\centering
\footnotesize
\caption{The results of the companion papers. ``Given'' lists the input columns of the table EvoMath searched, beyond
the physical parameter. ``Added after a miss'' names engine changes made after the engine failed on that target.}
\label{tab:cases}
\begin{tabular}{@{}p{3.3cm}p{2.6cm}p{5.1cm}p{3.6cm}@{}}
\toprule
Result & First obtained by & What EvoMath re-found, from what & Added after a miss \\
\midrule
Kerr shadow perimeter $6\sqrt3\,\pi\,{}_2F_1(-\tfrac16,\tfrac16;1;a^2)$, $M=1$ \cite{andrianov2026kerr} &
Hermite reduction and a contiguity relation & the ${}_2F_1$ from 40 rows of the spin alone; the prefactor came back as
the float 32.648388556216 & the ${}_2F_1$ operator family and seed \\
Kerr shadow area as a combination of elliptic periods \cite{andrianov2026kerr} & Hermite reduction & the integer
combination, given the periods as columns; the additive constant as a float ($-43.98$ for $-14\pi$) & the basis seed;
the link seed's late passes \\
Naked-cusp cross section $\sigma_3$ of the singular isothermal sphere with shear \cite{andrianov2026lens} & computer
algebra & the six-term law from 200 rows, given three angle variables of the construction & none \\
Finch's constant $I_1=(2/\pi)[3K(\tfrac14)-2E(\tfrac14)]$ \cite{andrianov2026lens} & PSLQ & not identified from 120
digits (a rational lookalike good to 5.5 digits); identified later by a seed & the elliptic constant seed \\
Radial dynamical-friction force on a circular orbit \cite{andrianov2026df} & least squares on a Weber-function basis,
then summation & the Clausen-function law from 65 rows, given the Kepler root $\theta_0=\mathcal M\sin\theta_0$ & the
Clausen operator and its columns \\
Subsonic eccentricity coefficient $\kappa(\mathcal M)$ \cite{andrianov2026df} & computer algebra & exact from 40
rows of $\mathcal M$ alone, default engine & two link-seed changes (ridge, $\operatorname{artanh}$ columns) \\
Supersonic $\kappa$ and precession coefficient $W$ \cite{andrianov2026df} & shell expansion & not found (best
$1-\Rsq$ of $2\times10^{-6}$ and $3\times10^{-8}$) & -- \\
Genus-two evaluation behind bilinear sequence regression \cite{andrianov2026bsr} & Hermite reduction and PSLQ &
nothing; no EvoMath run produced it & -- \\
NFW line-of-sight dispersion $J_1$ \cite{andrianov2026nfw} & derivation and computer algebra & the four-term
transcendental part, given the two weight-two functions as columns and with the elementary part subtracted from the
target; the full twelve-term law was missed, the aperture dispersion never found & none \\
\bottomrule
\end{tabular}
\end{table}

The same pattern holds across the other campaigns. With the elliptic-feature seed the engine recovered laws in
complete elliptic integrals from the physical parameter alone: the flat-band quantum metric of the staggered Lieb
lattice (10 of 10 runs, after three search passes were added for these laws; 4 of 9 before), the Kitaev honeycomb
energy at $J_x=J_y$ (8 of 8, with a seed written during that campaign and after a concurrency fix), linear spin-wave
constants, and the slip coefficient of an
elliptic duct (after the modulus map $1-x^2$ was added). From 120 digits the constant seeds identified lattice energies
at complex-multiplication points (after a period-Laurent shape was added) and box-integral constants as orthoscheme
volumes (with a seed designed from the solution). The algebraic seed recovered the Krein-collision curve of the
elliptic restricted three-body problem from data, which turned out to be in the literature
\cite{tschauner1971,matas1973}. The default engine, without opt-in seeds, recovered a handful of laws in these campaigns
(a precession rate in the Plummer potential, a view factor between equal spheres, the inductance of equal inclined
loops, a phase-curve law of a spheroid); on the campaign targets as a whole it found 4 of 22 in one block of rounds and
4 of 20 in the next, and none of 6 controls.

Three qualifications apply to almost every success:
\begin{enumerate}[leftmargin=1.5em,itemsep=1pt]
\item \emph{Designed inputs.} The tables often carried columns computed from the derivation: periods, angle variables,
Kepler roots, special-function values. With them the search reduces to finding integer or rational coefficients.
\item \emph{Changes after a miss.} Many seeds and passes were written after the engine failed on the target they then
solved. The successes show that the additions work; they do not show that the engine would have found the law
unaided.
\item \emph{Unidentified constants.} Overall scales and offsets were returned as fitted floats and recognised by hand
($6\sqrt3\pi$, $-14\pi$, $\gamma+2\ln2$). Identifying a constant is a separate step (\cref{sec:constants}).
\end{enumerate}

\paragraph{A record under the protocol.} The flat-band metric was later carried through the protocol of
\cref{sec:protocol} as a pilot dossier. Its record states the claim ``EvoMath confirmation, reviewed'', since the form
was known before recovery. The recovered formula is
$\mathcal M_{xx}(\delta)=[(1+\delta^2+\delta^4)K(m)-E(m)]/[4(1+\delta^2)]$ with $m=1-\delta^4$. It agrees with eight
holdout points, two outside the data range, to 39 digits; a second method (a Bloch Hamiltonian on a fine Brillouin-zone
grid) agrees to $2\times10^{-36}$; a fifteen-step derivation holds at every step to 40 digits; the validity audit passes
its eight checks, including the declared logarithmic singularity at $\delta\to0^+$ and the limits at both ends. Its proof
status is ``derivation, numerically checked'': none of the fifteen steps yet carries a symbolic proof. The independent
review scored blindness 3 of 5 and called the run a regression test of an engine tuned on the target, which is the
correct reading.

\section{Discussion}\label{sec:discussion}

\subsection{What the checks establish}

An ``exact'' verdict means that SymPy reduced the difference between the snapped model and a reference expression to
zero. It is a proof that two expressions are equal, under two conditions that are checked rather than assumed: the
translation reproduces the engine's predictions, and the snapped constants are the intended ones (a wrong snap makes
the symbolic step fail, it does not make it succeed). A failed symbolic step is not a disproof, since computer-algebra
simplification is incomplete; such a model is classed as equivalent or not recovered on numerical grounds.

The proof levels are an account of evidence, not a proof checker. Level 4 means that a human-readable derivation exists
and that each step was evaluated to 40 digits at seeded points; level 5 adds a SymPy verification of each step. Both can
miss an error that holds at the sampled points, and neither is a formal proof. No result in this paper has a formal
proof, and no proof assistant is connected.

The validity audit samples. It finds a pole or a branch cut between the points it evaluates, by bisection along each
input, and it compares declared limits with observed convergence orders. It does not prove that a formula is finite on
its whole domain, and in more than two inputs its grids are sparse.

\subsection{Threats to validity}\label{sec:threats}

\paragraph{The benchmark is ours.} The law suites were assembled while the engine was developed, and engine changes were
accepted only if they did not lose recoveries on them. EvoMath therefore has a home advantage that the comparison of
\cref{sec:compare} cannot remove: its structural templates, its constant packs (which include $\pi$, $e$, and physical
and medical constants) and its seeds are suited to laws of this kind, and the judge is its own verdict, applied with
the same leniency to all systems. Two rounds of changes (\cref{sec:lessons,sec:round2}) were made after EvoMath's misses
on these laws had been examined, and they account for 33 of its 180 exact recoveries, so part of its margin over the
other systems was fitted to this benchmark. v5 was written to test the
seed families on laws they had not been tuned on, and its four controls lie outside every family. A comparison on an external benchmark, such as the ground-truth problems of
SRBench \cite{lacava2021srbench} or SRBench++ \cite{defranca2024srbench2}, has not been run.

\paragraph{Competing interest.} The author is affiliated with EvoMath Research Labs UG, which offers EvoMath as
a commercial hosted service. The comparison should be read with this in mind; its data, harness, settings and
result files are published (see Code and data availability) so that it can be rerun independently.

\paragraph{Budget and settings.} Each system had four threads and 300\,s per law, and one random seed. Longer budgets,
several seeds or tuned settings may change the ranking. No system's settings were tuned on these suites.

\paragraph{Machine.} The runs shared a laptop (32 logical processors, 32\,GB of memory) with other work, and memory was
the binding constraint. Times are reported as measured but are indicative.

\paragraph{Code state.} Most of the engine described here was developed after the last tagged commit of version 1.
The tables were produced with the code of the repository tag \texttt{evomath-paper-v2}, except for the historical rows
and runs named as such; each result file of the comparison records a SHA-256 over the engine's modules.

\subsection{Limits of the engine}

Laws with nine to twelve terms on one-parameter data remain out of reach: their candidate families have numerical rank
14 to 19 in float64, and exhaustive support searches exceed any practical budget. The special-function seeds reach only
the moduli and arguments in their fixed lists; a modulus that carries lattice constants, an incomplete elliptic integral
of the third kind, or nested radicals are outside them. Recovery of special-function laws from the physical parameter
alone mostly needs the opt-in seeds; the default engine rarely finds them. Overall scales and offsets are fitted, not
identified. These are the places where further work would change what the system can do, and the discovery protocol
requires that such work widen a class, with tests on other members of the class and controls outside it.

\subsection{Changes from version 1}

This version corrects the first in several places. Its recovery table was illustrative; it is replaced by measured
results (\cref{sec:experiments}), and the scaling figure, which was not produced, is omitted. Several defaults were
stated wrongly or have changed: selection is $\varepsilon$-lexicase rather than tournament, constants are optimised
every 3 generations on 5 structures rather than every 25 on 10, the power exponent is clipped to $\pm10$, and
$\sinh$ and $\cosh$ are not in the default set. The argument for $\Rsq$ over Pearson's correlation applies only to
unscaled models (\cref{sec:fitness}). Simplification, description-length model selection and dimensional analysis,
listed as future work, now exist. The software licence is the PolyForm Noncommercial License 1.0.0, not a permissive
licence as stated before, and results are stored in SQLite by default, with PostgreSQL optional. The first version
also described a canonical-form method under the name IsalSR, attributed it to Bartlett et al.\ and gave the arXiv
number of a different paper: IsalSR is due to L\'opez-Rubio and Pascual-Gonz\'alez \cite{lopezrubio2026isalsr},
exhaustive symbolic regression to Bartlett et al.\ \cite{bartlett2023esr}, and arXiv:2109.13895 is the exhaustive
search of Kammerer et al.\ \cite{kammerer2020exhaustive}.

\section{Conclusion}\label{sec:conclusion}

EvoMath recovers exact laws from numerical tables when they lie in the reach of its evolutionary search or of its
seeds, and it keeps the evidence for each result apart: agreement on held-out points, symbolic identity, and a checked
derivation, with an audit of domains, branch cuts, singularities, limits and units, and a sealed record from which the
result can be repeated. On its own law suites, under an equal budget and judge, it recovered more laws exactly than PySR, Operon and gplearn, which fitted nearly as many laws but less often found the law itself; part of that margin comes from changes made after its misses on the same laws had been examined. In the
mathematical work behind five companion papers it confirmed closed forms that had been derived first, mostly with
derived quantities given as inputs or with seeds added after a miss, and it found none of them first. The protocol and
record described here are meant to make the next claim of that kind checkable by others.

\section*{Code and data availability}
The source code is at \url{https://gitlab.com/evomath-group/evomath}, under the PolyForm Noncommercial License 1.0.0
(source available, free for noncommercial use). A hosted version with a free tier is offered at \url{https://evomath.ai} by EvoMath Research Labs UG, with which the author is affiliated. The version used in this paper is the repository state tagged
\texttt{evomath-paper-v2}. The results of \cref{tab:standard,tab:constants,tab:compare} were produced with its engine
(commit \texttt{6ba0eb6}) and are committed under that tag; each result file of the comparison also records a SHA-256
over the engine modules it ran with, which equals that of the tagged engine. The historical rows and runs named as such
used the code of 20 November 2026 (commit \texttt{dca142b}) or the development tree of 4--7 November 2026; the
measurements that admitted the changes of \cref{sec:round2} ran from frozen copies of the development tree, and their
result files are committed as well. The comparison harness is in \texttt{backend/benchmarks/competitors}: the export
script with a manifest of SHA-256 hashes of every data file, the runners for the four systems, the scorer, and the
result files with each run's settings, environment and engine hash. The result files of
\cref{tab:standard,tab:constants,tab:compare}, the competitor settings and the list of the 216 laws with their
reference formulas are also provided as ancillary files with this preprint. The validity audit, the proof levels and
the discovery tool of \cref{sec:audit,sec:protocol,sec:record} are in \texttt{backend/evolmath/discovery}. The protocol
document and the pilot dossier of \cref{sec:cases}, with its sealed record, are not part of the public repository; the
author can provide them on request.

\section*{Acknowledgments}
The computations used NumPy, SciPy \cite{virtanen2020scipy}, SymPy \cite{meurer2017sympy}, mpmath \cite{mpmath} and
Matplotlib. The comparison used PySR, Operon and gplearn; we thank their authors for making them available. The AI
assistant Claude (Anthropic) was used, under the author's direction, to develop parts of the software, to run and
analyse the experiments, to derive and check results and to draft the text. The author is responsible for the
content.

\bibliographystyle{unsrt}
\bibliography{refs}

\appendix
\section{Configuration}\label{app:config}

\begin{table}[h]
\centering
\small
\caption{Engine settings: the defaults of \texttt{GPConfig}, the law-suite configuration of \cref{sec:standard}, and
the comparison runs of \cref{sec:compare} (which use the suite configuration with two overrides).}
\label{tab:defaults}
\begin{tabular}{@{}p{4.6cm}p{5.6cm}p{4.6cm}@{}}
\toprule
Setting & Default & Law suites \\
\midrule
population per island, islands & 500, 4 & 300, 4 \\
maximum generations, maximum depth & 5000, 6 & 120, 7 \\
selection & $\varepsilon$-lexicase, 128 cases & same \\
crossover, mutation, elitism & 0.9, 0.1, 5 & same \\
migration & ring, every 25 generations, 3 migrants & same \\
cross-validation folds, holdout, screening rows & 3, 15\%, 200 & same \\
parsimony per node & 0.005 & same \\
constant optimisation & L-BFGS-B, every 3 generations, top 5 structures & same \\
early stop (holdout $\Rsq$) & 0.9999 & 0.999 \\
stall rule, restarts & off, 0 & 3 reports, 2 \\
residual boosting & 1 round, child at most 200 generations & same \\
seeds on by default & sparse 4, link 6, Laurent 1; box (one search per run, shared by the islands) & same \\
noise-band snapping of the final model & on & same \\
wall-clock limit, evaluation workers & none, automatic & none, automatic (comparison: 300\,s, 4) \\
random seed & 42 & 42 \\
\bottomrule
\end{tabular}
\end{table}

\begin{table}[h]
\centering
\small
\caption{Settings of the other systems in the comparison. All had four threads, a 300\,s limit per law and random seed
42. PySR and gplearn stopped early when the training error reached rounding level; Operon ran its own budget.}
\label{tab:competitor-settings}
\begin{tabular}{@{}lp{11.5cm}@{}}
\toprule
System & Settings \\
\midrule
PySR & binary $+,-,\times,\div,\wedge$; unary $\exp,\log,\sqrt{\cdot},\sin,\cos$; maximum size 30; constraint
$\wedge:(-1,1)$ (exponent a constant or variable); otherwise defaults; multithreaded (not bitwise reproducible) \\
Operon & symbols $+,-,\times,\div$, constant, variable, $\exp,\log,\sqrt{\cdot}$, power, $\sin,\cos$; maximum length
50; 1000 generations, $10^6$ evaluations, 10 Levenberg--Marquardt iterations per individual; model selection by
description length; otherwise defaults \\
gplearn & population 1000, tournament 20; functions $+,-,\times,\div,\sqrt{\cdot},\log$, negation, reciprocal,
$\sin,\cos$ and a protected $\exp$; mean squared error; parsimony $10^{-3}\operatorname{var}(y)$ per node (a fixed
$10^{-3}$ let programs grow until one generation took minutes, and the automatic setting collapsed to a single input);
constants in $[-5,5]$; continued one generation at a time until the limit \\
\bottomrule
\end{tabular}
\end{table}

\end{document}